\documentclass{article}
\usepackage[final]{corl_2019} % uncomment for the camera-ready ``final'' version

\usepackage{times}
\usepackage{multicol}
\usepackage{hyperref}
\usepackage{graphicx}
\usepackage{bm}
\usepackage[nolist]{acronym}
\usepackage{comment}
\usepackage{amsmath}
\usepackage{amssymb}
\usepackage[capitalise]{cleveref}
\usepackage{algorithm}
\usepackage[usenames,dvipsnames]{xcolor}
\usepackage{algcompatible}
\usepackage{ulem}
\usepackage[caption=false,font=footnotesize]{subfig}
\usepackage{booktabs}
\usepackage{array}
\usepackage{dblfloatfix}
\usepackage{wrapfig}
\usepackage{tabularx}
\usepackage{enumitem}

\newcommand{\norm}[1]{\left\lVert#1\right\rVert}

\newcommand{\bracket}[1]{\left[ #1 \right]}

\begin{document}

% paper title
\title{Perceptual Attention-based Predictive Control}

% \author{
%   Keuntaek~Lee, Gabriel Nakajima~An, Viacheslav~Zakharov, and Evangelos A.~Theodorou\\
%   Autonomous Control and Decision Systems Laboratory\\
%   Georgia Institute of Technology\\
%   \texttt{keuntaek.lee@gatech.edu}\\
% }

\author{
  Keuntaek Lee\\
  Georgia Institute of Technology\\
  \texttt{keuntaek.lee@gatech.edu} \\
  \And
  Gabriel Nakajima~An \\
  Georgia Institute of Technology \\
  \texttt{gabriel.nakajima.an@gmail.com } \\
  \And
  Viacheslav~Zakharov \\
  Georgia Institute of Technology \\
  \texttt{vzakharov3@gatech.edu } \\
  \And
  Evangelos A. Theodorou \\
  Georgia Institute of Technology \\
  \texttt{evangelos.theodorou@gatech.edu }
}

\begin{acronym}
\acro{NN}{Neural Network}
\acro{CNN}{Convolutional Neural Network}
\acro{DL}{Deep Learning}
\acro{hl}{\textit{heteroscedastic} loss}
\acro{KL}{Kullback-Leibler}
\acro{IL}{Imitation Learning}
\acro{iLQG/MPC-DDP}{iterative Linear Quadratic Gaussian/Model Predictive Control Differential Dynamic Programming}
\acro{MPC}{Model Predictive Control}
\acro{DDP}{Differential Dynamic Programming}
\acro{BNN}{Bayesian Neural Network}
\acro{MP-Net}{Model Prediction Network}
\acro{Macula-Net}{Macula-Network}
\acro{PAPC}{Perceptual Attention-based Predictive Control}
\acro{ROI}{Region of Interest}
\end{acronym}

\maketitle

\begin{abstract}
In this paper, we present a novel information processing architecture for \textit{safe} deep learning-based visual navigation of autonomous systems. The proposed information processing architecture is used to support a perceptual \textit{attention-based} predictive control algorithm that leverages model predictive control (MPC), convolutional neural networks (CNNs), and \textit{uncertainty quantification} methods. The novelty of our approach lies in using MPC to learn how to place attention on relevant areas of the visual input, which ultimately allows the system to more rapidly detect unsafe conditions. We accomplish this by using MPC to learn to select \textit{regions of interest} in the input image, which are used to output control actions as well as estimates of \textit{epistemic} and \textit{aleatoric uncertainty} in the attention-aware visual input. We use these uncertainty estimates to quantify the safety of our network controller under the current navigation condition. The proposed architecture and algorithm is tested on a 1:5 scale terrestrial vehicle. Experimental results show that the proposed algorithm outperforms previous approaches on early detection of unsafe conditions, such as when novel obstacles are present in the navigation environment. The proposed architecture is the first step towards using deep learning-based perceptual control policies in \textit{safety-critical} domains.
\end{abstract}

\keywords{Safe Imitation Learning, Anomaly Detection, Uncertainty Quantification, Bayesian Neural Networks, Perceptual Control, MPC, Autonomous Driving\\
\textbf{Supplementary video}: \url{https://youtu.be/-Zmi0HCvM9I}
}

\section{Introduction}

For autonomous systems to be able to operate in uncertain environments, they have to be equipped with robust decision-making capabilities using a variety of perceptual modalities including vision.  Recent advancements in Artificial Intelligence and Deep Learning,  have facilitated the development of algorithms that integrate perception and control in a holistic fashion. The resulting \textit{perceptual control policies} offer unique capabilities with respect to generalization, representation, and performance in tasks such as vision-based navigation.  

Prior work on vision-based navigation mostly relies on object detection and segmentation. Recently, \citet{fcn, fasterrcnn, maskrcnn} showed improved performance in instance segmentation where both object detection and semantic segmentation is handled. Proceeding this vision step, path planning/trajectory optimization is performed followed by control.

Alternative methodologies for performing vision-based navigation are via \ac{IL}, also referred to as learning from demonstration. In the \ac{IL} framework, the learning algorithm has access to an expert policy to take advantage from. This expert policy can come, for instance, from human demonstrations or \ac{MPC}. For the autonomous driving task, \citet{bojarski2016end} proposed an approach for learning to drive a full-size car autonomously directly from vision data. Moreover, \citet{PanRSS18} accomplished high-speed autonomous driving via an end-to-end \ac{IL} approach using DAgger \citep{DAgger} algorithm.

Nonetheless, the tremendous success of these methods cannot diminish the importance of \textit{safety}. In particular, safety is a crucial topic in domains with long-tailed input distributions, such as in navigation where novel objects can appear in the visual input. Our previous work \citep{Lee2019EarlyFailure, Lee2019Ensemble} addressed the problem of incorporating safety into end-to-end trained controllers via \acp{BNN}. By using \ac{BNN}'s ability to output distributions of its predictions, we were able to quantify the \textit{uncertainty} in the control output of our network policy. When encountering unseen data in the long tails, our \ac{BNN} policy would output high-variance (high uncertainty) distributions. This, in turn, enabled the system to pass the control authority to a safe expert before the end-to-end trained controller failed.

Despite the success of this approach, the system was appreciably slow at detecting such an increase in uncertainty in the control output. In particular, when encountering new obstacles, the system was only able to detect high uncertainty when reaching too close to the object. This consequently impeded the control authority to be passed to a safe expert in sufficient time, which resulted in aggressive maneuvers and ultimately accidents. This limitation was a consequence of the model's inability to detect high uncertainty when such anomaly obstacles were distant and occupying limited space in the visual input.

In this work, we address this major limitation by viewing perceptual control policies as Information Processing Architectures (IPAs) and proposing a novel algorithm for safe vision-based control with quick anomaly detection. We accomplish this by introducing the \ac{PAPC} algorithm, which is comprised of the following ingredients: (1) it uses \ac{IL} to learn a perceptual controller; (2) it employs \acp{BNN} to estimate the uncertainty of the learned controller; and (3) it incorporates a novel \textit{attention mechanism} that enhances the uncertainty quantification method to be able to promptly assess unsafe navigation conditions.

% In the \ac{PAPC} algorithm, decision-making and perception are tightly coupled since predicted state trajectories force perception to focus on relevant areas of the input visual information. This attention mechanism enables early detection of unseen situations, such as the cases when a new obstacle appears in the driving lane. When such situations arise, the network policy is to concede control to a safe policy or expert, such as a fully observable MPC controller. \todo{make this thing extremely clear}
In particular, our attention mechanism takes advantage that our \ac{MPC} expert policy outputs control and \textit{state trajectories}. We process these trajectories to train a model to predict the vehicle's future trajectory in the pixel space of the visual input. Finally, these pixel trajectories are then used to place attention on certain parts of the visual input, ultimately enabling our algorithm to detect unsafe conditions significantly quicker. In this work, we simulate safety hazards with the presence of unseen obstacles in the navigation path.

In summary, the contributions of this work are provided as follows: 
\begin{itemize}[leftmargin=0.4cm]
  \item We introduce the Model Prediction-Network (MP-Net) for learning the vehicle's trajectories represented as splines (\textit{spline learning}) in pixel space. The MP-Net is trained using state trajectories generated by \ac{MPC}. We use MP-Net's output splines to select \acp{ROI} in the visual input, where attention is to be placed on.
  
  \item We present the eye-inspired Macula-Net, a 3D \ac{CNN} that uses the \acp{ROI} mentioned above as input and generates controls as well as estimates of aleatoric and epistemic uncertainty.
%   The Macula-Net is trained in a Bayesian fashion using input-output pairs consisting of ROIs and corresponding control commands generated by MPC.
  
  \item We integrate all the aforementioned blocks into the \ac{PAPC} algorithm. PAPC outperforms the prior state-of-the-art solutions on assessing unsafe navigation conditions via early detection of novel obstacles in the vehicle's way. In contrast to traditional methods, detection of these obstacles is performed without any image classification or object detection.
%   \item The PAPC algorithm is tested and compared against prior state-of-the-art solutions. Experiments are performed in simulation as well as on real hardware and demonstrate the benefits and outperformance of PAPC.
\end{itemize}

% The remaining of the paper is organized as follows: In \cref{sec:Preliminaries}, we briefly review some preliminaries used in our work. In \cref{sec:Model_Prediction_Network}, we introduce the Model Prediction Network, which predicts the future location of the vehicle in pixel coordinates used to construct ROI windows. In \cref{sec:Perceptual_Attention}, we introduce the Macula-Net, which processes the ROIs and outputs a control mean and variance. We also detail our Perceptual Attention-based Predictive Control (PAPC) algorithm. \cref{sec:Experiments} details simulation and real hardware experiments with analysis and comparisons of the proposed methods. Finally, we conclude and discuss future directions in \cref{sec:discussion} and \cref{sec:conclusion}.

\section{Preliminaries}
\label{sec:Preliminaries}

In this section, we detail the building blocks of the proposed Information Processing Architecture (IPA) for perceptual control.

\subsection{Model Predictive Optimal Control}
\ac{MPC}-based optimal controllers (e.g. \ac{iLQG/MPC-DDP} \citep{MPCDDP}, Model Predictive Path Integral \citep{mppi}) provide planned control trajectories by solving the optimal control problem within a time horizon. An optimal control problem whose objective is to minimize a task-specific cost function can be formulated as follows:
\begin{align}\label{eq:optcontrolprob}
	V(x(t_0), t_0) = \min_{u(t)}\bracket{\phi(x(t_f), t_f) + \int_{t_0}^{t_f}\ell(x(t), u(t), t) dt}
\end{align}
subject to dynamics $\frac{dx}{dt} = f(x(t), u(t), t)$, where $x \in \mathbb{R}^n$ represents the system states, $u \in \mathcal{U} \subset \mathbb{R}^m$ represents the control, $\phi$ is the state cost at the final time $t_f$, $\ell$ is the running cost, and $V$ is the value function. 
By solving this optimization problem, we get the future optimal state trajectories from the optimal control trajectories.

In this paper, we take advantage of the optimal state and control trajectories provided by MPC to train perceptual control policies in an \ac{IL} fashion and design an attention mechanism. As will be explained later, the state trajectories will be used to train CNNs to predict ROIs using raw images while control action will be used to train another CNN to predict the control action using the \acp{ROI} as input.

\subsection{Perceptual Control via Imitation Learning} \label{sec:il}

One of the deep learning-based perceptual control approaches for a navigation task is training agents to output optimal control actions given image data from cameras.

Reinforcement Learning (RL) is one way to train agents to maximize some notion of task-specific rewards. One of the major problems in RL is the sample-inefficiency problem: the agents have to randomly explore the action-state space without any prior knowledge of the environment or task.

However, IL uses supervised learning to train a control policy and bypasses the sample-inefficiency problem in RL. In IL, a policy is trained to accomplish a specific task by mimicking an expert's control policy, which in most cases, is assumed to be optimal. Accordingly, IL provides a safer training process. In this work, we train our control policy in an IL fashion and we use MPC as the expert policy.

One of the major problems in IL is that the training data collected from an optimal expert does not usually include demonstrations of failure cases in unsafe situations. \citet{DAgger} introduced an online dataset aggregation algorithm, DAgger, which mixes the expert's policy and the learner's policy to explore various situations. However, even with the online scheme of collecting datasets, it is impossible to experience all kinds of unexpected scenarios.
Therefore, in order to deploy the trained policy into safety-critical systems, we need a strategy to detect when our trained model is going to fail or provide unsafe results.
%The idea of quantifying the uncertainty in the model output using Bayesian Neural networks will be introduced in the following section.

% \ac{IL} is one approach to learn how to do a specific task by imitating a teacher's or an expert's control policy. In IL settings, it is usually assumed that the expert is perfect and always makes optimal decisions. The IL framework allows us to do end-to-end control from raw sensor input since it bypasses all the burdensome steps in navigation (perception, filtering, localization, path planning, etc.) and directly applies control only with given observations.

% To learn by imitating an expert, the goal of IL becomes learning a policy that minimizes the difference between the task-specific cost (\cref{eq:optcontrolprob}) compounded by the expert's policy and that incurred by the learner. To achieve the goal, the learned policy should aim to converge to the expert's policy.
% For a deep learning based IL, the mean squared error between the network's predictions and the ground truth control actions labeled by an expert is widely used to train the learner network.

\subsection{Bayesian Neural Networks}
 
The quantification of uncertainty in the model output is a crucial component for the deployment of DNNs to safety-critical applications. To incorporate this capability to deep learning-based perceptual control policies, we use \acl{BNN}s.
Currently, Bayes by backpropagation \citep{pmlr-v37-blundell15}, Monte Carlo (MC) dropout \citep{pmlr-v48-gal16}, and Deep Ensembles \citep{Lakshminarayanan_17_deepensemble} are the most common methods to instantiate \acp{BNN}. In our work, we use the MC-dropout method, which uses the dropout technique to build a probability distribution over network weights. This allows us to obtain the distribution of the network prediction at test time.

% This output distribution comes from the input distribution, where the trained Bayesian network will output a large output variance if the input distribution at the test time is largely different from the training input distribution.

\citet{WhatUncertainties} introduced the \textit{heteroscedastic} loss function which provides two different notions of uncertainty: aleatoric and epistemic. Aleatoric uncertainty originates from incomplete knowledge of the environment whereas epistemic uncertainty arises from the lack of sufficient data.

In this work, we use this \textit{heteroscedastic} loss function to train our Bayesian network and we use the network's output variance for the early assessment of unsafe conditions simulated by novel obstacles in the vehicle's way.

\subsection{B-spline} \label{sec:B-spline}
The B-spline is a collection of Bezier splines that are defined by a set of knot coordinates around which each spline is centered \citep{Bspline}. This set of splines has the following continuity requirements: i) The end of the previous curve must have the same value as the start of the next. ii) The first and second derivatives must be conserved between the intersecting points.
Therefore, a high-degree B-spline can smoothly approximate a curve. The equation for a k-degree B-Spline is formulated as $S(t) = \sum_{i=0}^n N_{i,k} (t)P_i,$ where ($P_0, P_1, \ldots, P_n$) are control points and $N_{i,k}(t)$ are the basis functions defined using the recursive Cox-de Boor formula \citep{cdboorformula}. 
% as follows:
% \begin{align*}
    % N_{i,j}(t) = \frac{t-t_i}{t_{i+j} - t_i} N_{i,j-1}(t)+ \frac{t_{i+j+1}-t}{t_{i+j+1}-t_(i+1)} N_{i+1,j-1}(t)
% \end{align*}
% where $t_i$ is a knot vector. Knot vectors determine which of the basis functions governs the spline curve at a particular point.
In this work, the B-spline coefficients were used to train the Model Prediction Network described in the next section.

\section{Model Prediction Network}

Our attention mechanism revolves around the idea that the model should focus on areas (ROIs) of the visual input that the vehicle will navigate towards. Fortunately, the expert MPC's future state trajectory output provides us with exactly this vehicle course information. 

\begin{figure}[h]
    \vspace{-0.1cm}
    \centering
    \includegraphics[width=0.99\textwidth]{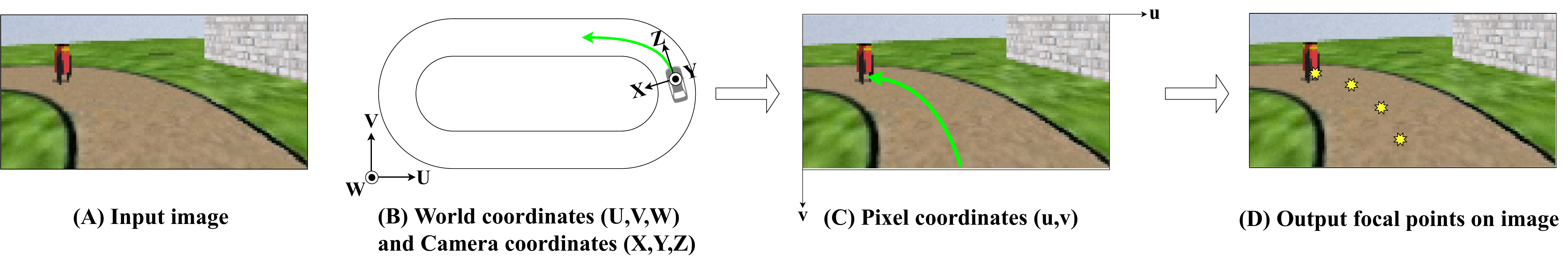}
    \caption{\footnotesize An overview of the Model Prediction Network (MP-Net). (B) From IMU/GPS data, the system states are estimated and the model predictive controller generates the model's future path/position in the world coordinates. From (A) the input image, the \ac{MP-Net} is trained to predict (C) the B-spline coefficients of the predicted state trajectory in the pixel coordinates. From the spline coefficients predicted by the \ac{MP-Net}, we reconstruct the spline and (D) choose focal points from the reconstructed spline. }\label{fig:MP}
    \vspace{-0.2cm}
\end{figure}

We first transform the MPC's state trajectory in the original state space to a corresponding trajectory in pixel coordinates represented by a spline (as in \cref{fig:MP} (C)). Next, we use these spline trajectories as targets to train a CNN model we refer to as Model Prediction Network (MP-Net). At test time, when there is no access to the MPC expert, the MP-Net will output a spline trajectory in pixel space, given the vehicle's visual input. Subsequently, we select a specific number of \textit{focal points} (as in \cref{fig:MP} (D)) along the outputted pixel trajectory, and use these points to create the \ac{ROI} windows.
This mechanism allows the system to put attention on the image regions corresponding to the vehicle's trajectory, therefore increasing sensitivity to safety hazards along this trajectory, even if such hazards are located far away and rendered small in the visual input.
% an attention mechanism allows our safe policy to be sensitive even to unseen obstacles that are far away and small in its image view, as long as they are in the way of its trajectory.

% The mapping of the state trajectory from state space to pixel coordinates will be implicitly learned with a deep convolutional neural network (CNN) we refer to as Model Prediction Network (MP-Net).
% \ac{MP-Net} has a similar network structure as VGG16 \citep{VGG}, a widely used deep convolutional neural network structure that learns the mapping between input images and the corresponding class labels, and the network is trained with the mean squared error loss.
% We collect training data from \ac{MPC}, which provides planned control trajectories as well as state trajectories at every time step. We set the B-spline coefficients of the \ac{MPC}-planned state trajectories in the pixel coordinate as targets.
% This network will take an image as input and will output spline coefficients of a trajectory in pixel coordinates.
In \cref{sec:coordinate_transform}, we describe how we obtain the target spline trajectories to train the MP-Net, and in \cref{sec:training_mp_net}, we detail how the MP-Net is trained on those targets.

\label{sec:Model_Prediction_Network}

% The \ac{MPC}-based controllers provide the future state (e.g. position, velocity) trajectories. Inspired by \ac{MPC}, we introduce a network that predicts a robot's future positions in the image space. Based on the predicted future trajectories, we find \acp{ROI} that the robot can put attention on.
% \todo{talk about key idea, which is the common sense part}
% This allows us to ``see'' the path (in the image) that the vehicle will move along.

\subsection{Targets for MP-Net using Coordinate Transformation}
\label{sec:coordinate_transform}
As seen in \cref{fig:MP}, \ac{MP-Net} projects the vehicle's future state trajectory described in the world coordinates onto a 2D image in a moving frame of reference. This coordinate transformation technique is widely used in 3D computer graphics \citep{coordtransformbook}.
The coordinate transformation consists of 4 steps: $\text{World} \rightarrow \text{Robot} \rightarrow \text{Camera} \rightarrow \text{Film} \rightarrow \text{Pixel}.$
We follow the convention in the computer graphics community  and set: the \textbf{Z} (optic)-axis as the vehicle's longitudinal (roll) axis; the \textbf{Y}-axis as the axis normal to the road, where the positive direction points upwards; and the \textbf{X}-axis as one perpendicular on the vehicle's longitudinal axis, where the positive direction points to the left of the vehicle.

Let us define roll, pitch, yaw angles as $\phi, \theta, \psi$, respectively, the camera (vehicle) position as $U_{\text{cam}}, V_{\text{cam}}, W_{\text{cam}}$ in world coordinates, and the camera focal length as $f$.
% Additionally, we subtract (0.5 m) from this depth. This is done to account for a discrepancy between the camera location at the front of the vehicle and the local reference frame origin at the center of the vehicle.
Then, we construct the rotation matrices around the \textbf{U}, \textbf{V}, \textbf{W}-axis $R_U, R_V, R_W$, the translation matrix $T_{\text{tl}}$, the robot-to-camera coordinate transformation matrix $T_{r \rightarrow c}$, and the projection matrix $T_{c \rightarrow f \rightarrow p}$ as:
\\
\begin{align*}
R_U =
\begin{bmatrix}
    1 & 0 & 0 \\
    0 & cos\phi & -sin\phi \\
    0 & sin\phi & cos\phi
\end{bmatrix}, 
R_V =
\begin{bmatrix}
    cos\theta & 0 & sin\theta \\
    0 & 1 & 0 \\
    -sin\theta & 0 & cos\theta
\end{bmatrix},
R_W =
\begin{bmatrix}
    cos\psi & -sin\psi & 0 \\
    sin\psi & cos\psi & 0 \\
    0 & 0 & 1
\end{bmatrix},
\\
T_{\text{tl}} =
\begin{bmatrix}
    1 & 0 & 0 & -U_{\text{cam}} \\
    0 & 1 & 0 & -V_{\text{cam}} \\
    0 & 0 & 1 & -W_{\text{cam}}
\end{bmatrix},
T_{r \rightarrow c} =
\begin{bmatrix}
    0 & 1 & 0 \\
    0 & 0 & 1 \\
    1 & 0 & 0
\end{bmatrix},
T_{c \rightarrow f \rightarrow p} =
\begin{bmatrix}
    f/Z & 0 & o_X/Z \\
    0 & f/Z & o_Y/Z
\end{bmatrix},
\end{align*}
where the projection matrix  $T_{c \rightarrow f \rightarrow p}$ projects the point $(X, Y, Z)$ in the camera coordinates into the film coordinates using the perspective projection equations \citep{coordtransformbook} and the offsets $o_X$ and $o_Y$ transform the film coordinates to the pixel coordinates by shifting the origin.

In addition, the total rotation matrix $R$, the world-to-robot coordinate transformation matrix $T_{w \rightarrow r}$, and the world-to-pixel transformation matrix $T$ are calculated as
\begin{align*}
     R = R_W R_V R_U \in \mathbb{R}^{3\times3}, \hspace{0.5cm} T_{w \rightarrow r} = RT_{\text{tl}} \in \mathbb{R}^{3\times4}, \hspace{0.5cm} T = T_{c \rightarrow f \rightarrow p}T_{r \rightarrow c}T_{w \rightarrow r}  \in \mathbb{R}^{2\times4}.
\end{align*}
After converting the \textbf{X}, \textbf{Y}, \textbf{Z}-axes to follow the convention in the computer vision community through $T_{r \rightarrow c}$, the projection matrix $T_{c \rightarrow f \rightarrow p}$ converts the camera coordinates to the pixel coordinates.

We obtain the vehicle (camera) position in pixel coordinates (u,v) with $[\text{u}', \text{v}']^T = T [U_{cam}, V_{cam}, W_{cam}, 1]^T$.
% \begin{align}
%     \begin{bmatrix}
%         u' \\ v'
%     \end{bmatrix}
%     =
%     T
%     \begin{bmatrix}
%         U_{cam} \\ V_{cam} \\ W_{cam} \\ 1
%     \end{bmatrix}.
% \end{align}
However, this coordinate-transformed point $[\text{u}', \text{v}']$ in the pixel coordinates has the origin at the top left corner of the image. In our work, as we deal with the state trajectory of the vehicle, we rotate the axes by switching u$'$ and v$'$, and we define the new origin at the bottom center of the image $[\frac{w}{2}, h]$, where $h$ and $w$ represent the height and width of the image, respectively. Finally, we subtract $[\text{v}', \text{u}']$ from $[\frac{w}{2}, h]$ and get the final expressions $[\text{u}, \text{v}] = [\frac{w}{2}, h] - [\text{v}', \text{u}']$.
% \begin{align}
%     \begin{bmatrix}
%         u \\ v
%     \end{bmatrix}
%     =
%     \begin{bmatrix}
%         w/2 \\ h
%     \end{bmatrix}
%     -
%     \begin{bmatrix}
%         v' \\ u'
%     \end{bmatrix}
% .
% \end{align}

\subsection{Training MP-Net} \label{sec:training_mp_net}
Instead of training the \ac{MP-Net} to predict the entire trajectory in the pixel coordinates, we train it to learn the spline coefficients of the trajectory. This is possible because the MPC trajectories are simple and smooth enough to be represented with splines. This greatly simplifies the regression problem, without jeopardizing performance. To train for spline coefficients, we first fit a spline through the vehicle's pixel trajectory and we regress on the spline coefficients. At test time, when we don't have access to the MPC expert, we predict spline coefficients and sample a fixed number of \textit{focal points} along this spline to create the ROIs.

Another way to generate the focal points is by \textit{directly} regressing them in pixel space. However, this is not flexible to changes in the number of focal points, as the network would have to be re-trained for a different number of points. Our \textit{spline-learning} approach allows us to generate any number of focal points, which proved to be very useful during experimentation.

We compared the prediction error of the \textit{spline-learning} and the \textit{direct} focal points learning method. For a fair comparison, we used the same CNN architecture for both methods.
For \textit{spline-learning}, we trained the MP-Net to predict the eight B-spline coefficients and for the \textit{direct} focal points learning, we trained the same model to predict the four focal points $[u, v]$ in the pixel coordinates. Our experiments showed that the \textit{spline-learning} method required much fewer training epochs and it clearly outperformed the \textit{direct} focal points learning approach.
% which was trained with 100 times more training epochs.
The average testing error (MSE) in pixel space was 0.4 for the \textit{spline-learning} method and 25.2 for the \textit{direct} focal points learning method. Here, we argue that even with the same number of values, the spline coefficients carry much more information than pixel position values do.

\begin{figure*}[h]
    \vspace{-0.2cm}
    \centering
    \includegraphics[width=0.99\textwidth]{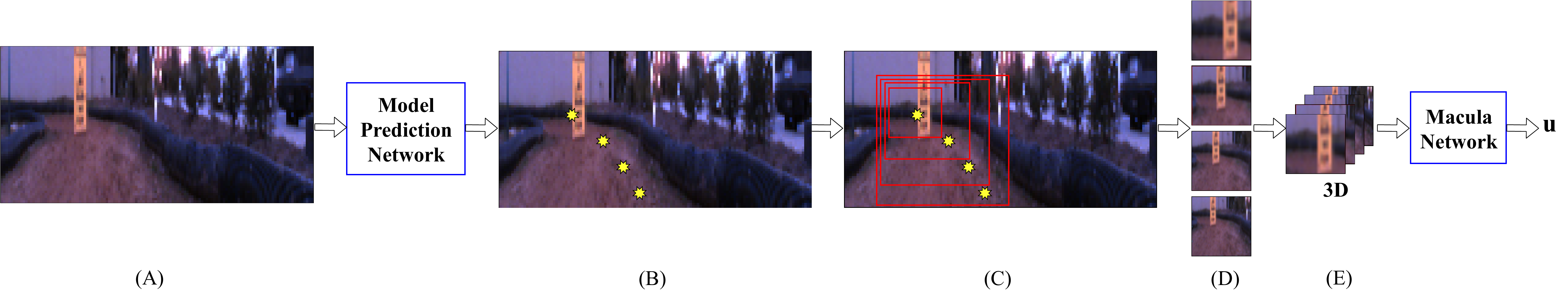}
    \caption{\footnotesize An overview of the Perceptual Attention-based Predictive Control (PAPC) algorithm. (A) Input RGB image with size (64, 128, 3). (B) Focal points (yellow) from the \ac{MP-Net}-predicted spline. (C) Constructed \acp{ROI} from the focal points. (D) Resized \acp{ROI} with the same size (32, 32, 3). Bigger \ac{ROI} loses more resolution by downsampling. (E) Stacked 2D images into 3D data (4, 32, 32, 3). This multi-resolution 3D data resembles the input to the macula in human eyes. }
    \label{fig:MPM}
\end{figure*}

\begin{wrapfigure}{R}{0.55\textwidth}
\vspace{-0.7cm}
\centering
\includegraphics[width=0.55\textwidth]{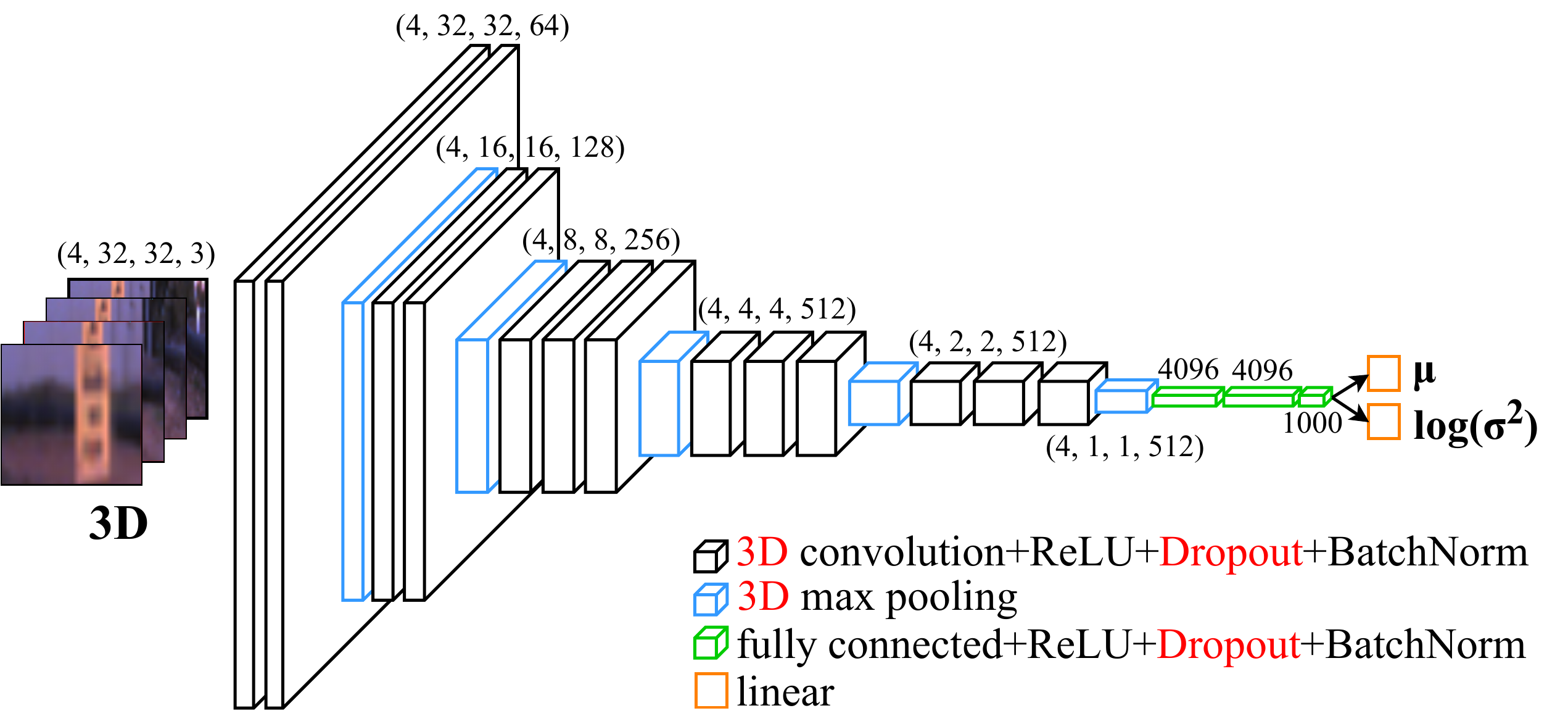}
\caption{\footnotesize The Macula-Network structure having 3D image data as an input and control action (mean and variance) as an output. The network is the 3D version of VGG16 Network \citep{VGG} with a Bayesian scheme by dropout.}
\vspace{-0.7cm}
\label{fig:Macula}
\end{wrapfigure}

% \vspace{-0.4cm}
\section{Perceptual Attention-based Predictive Control}
\label{sec:Perceptual_Attention}

As described in \cref{fig:MPM}, once the focal points are obtained from the MP-Net-predicted splines, we (1) construct ROI windows according to these focal points, (2) perform downsampling to lose some resolution, which results in putting more attention on the less-downsampled regions, and (3) feed these processed ROIs into the Macula-Net. The Macula-Net is named after the central part of our eyes' retinas, where we get the clearest vision with most resolution. The Macula-Net will take these multi-resolution ROIs and output a control mean and variance via the Bayesian MC-dropout method \citep{pmlr-v48-gal16}.
For the architecture of the Macula-Net (\cref{fig:Macula}), we adopted the 3D version of VGG 16 \citep{VGG}.
% , even though researchers have been recently developed smaller network structures with better accuracy. This is because of the easiness of the VGG structure, where we can simply apply the Concrete Dropout \citep{Gal2017Concrete} method on top of it.
%Although the network size is not small, compared to other popular CNN structures, we are still able to get a good approximation of the Bayesian Network via Monte Carlo sampling with around 25 samples in real-time (20Hz). \todo{KT, i need you to better explain this}
In addition, to run the PAPC algorithm in real-time (20Hz), 25 Monte Carlo samples were used in the Macula-Net.

The Macula-Net is trained using the \textit{heteroscedastic} loss function \citep{WhatUncertainties} to produce a distribution over control actions as an output. This loss function is defined as
\begin{equation} \label{eq: loss function}
    \mathcal{L}_h(y, \mu, \sigma) = \sum_i \frac{\norm{{y_i - \mu}}_2^2}{\sigma^2} + \log \sigma^2,
\end{equation}
where $y$ is the target data, and $\mu$ and $\sigma$ represent the Gaussian distribution of the prediction.

The ROIs are constructed in 3 steps: First, define the fovea focal point $p_{\text{fovea}}$ as the farthest focal point along the spline. Second, construct the smallest \ac{ROI}, referred to as fovea, as a window of size 32x32 with $p_{\text{fovea}}$ as its center. Third, for each of the other focal points $p_{i}$, construct an ROI with center $p_{\text{mid}} = (p_{i}+p_{\text{fovea}})/2$ and a window size of ($w_u, w_v$), where $w_u = 2(p_{\text{mid},u}-p_{\text{fovea},u}) + \textit{margin}, w_v = 2(p_{\text{mid},v}-p_{\text{fovea},v}) + \textit{margin}.$
In this way, each \ac{ROI} can cover the corresponding focal point and the fovea with some \textit{margin}, defined manually as a hyperparameter.

\begin{wrapfigure}{R}{0.62\textwidth}
\vspace{-0.3cm}
\begin{minipage}{0.62\textwidth}
\begin{algorithm}[H]
\footnotesize
\caption{Perceptual Attention-based Predictive Control (PAPC)}
\begin{algorithmic}[1]
\label{alg:PAPC}
    \REQUIRE $\newline \mathbf{o}_{\text{img},t} \text{: Image from a camera at timestep $t$, } \newline \mathbf{o}_{\text{world}, t} \text{: States in world coordinates at timestep $t$, } \newline \mathbf{N} \text{: Number of data points for training, } \hspace{0.2cm} \mathbf{T} \text{: MPC timesteps }$
\FOR{$t = 0 : \mathbf{N}$}
\STATE $u_{t,...,t+T}, [U, V, W, \phi, \theta, \psi]_{t,...,t+T} \leftarrow $MPC$(\mathbf{o}_{\text{world},t})$
\ENDFOR
\STATEx \textcolor{gray}{$\%$ Generate a dataset to train MP-Net. }
\STATE $[\text{u}, \text{v}] \leftarrow $ CoordinateTransform$(U, V, W, \phi, \theta, \psi)$ \textcolor{gray}{; \cref{sec:coordinate_transform}}
\STATE $C_{\text{pixel}} \leftarrow $ Spline$(\text{u}, \text{v})$ \textcolor{gray}{; \cref{sec:B-spline}}
\STATE $\text{ROIs} \leftarrow $ GenerateROIs$(\mathbf{o}_{\text{img}}, C_{\text{pixel}})$ \textcolor{gray}{; \cref{sec:Perceptual_Attention}}
\WHILE{Training MP-Net}
\STATE $\hat{C}_{\text{pixel}} \leftarrow $ MP-Net$(\mathbf{o}_{\text{img}})$
\STATE $\text{Loss}=\text{MSE}(C_{\text{pixel}}, \hat{C}_{\text{pixel}})$
\STATE Update MP-Net
\ENDWHILE
\WHILE{Training Macula-Net}
\STATE $\text{ROIs} \leftarrow  \text{GenerateROIs}\big(\mathbf{o}_{\text{img}}, \text{MP-Net}(\mathbf{o}_{\text{img}})\big)$
\STATE $\hat{u}_{\text{mean}}, \hat{u}_{\text{var}} \leftarrow $ Macula-Net$(\text{ROIs})$
\STATE $\text{Loss}=\mathcal{L}_h(u, \hat{u}_{\text{mean}}, \hat{u}_{\text{var}})$  \textcolor{gray}{; \cref{eq: loss function}}
\STATE Update Macula-Net
\ENDWHILE
\WHILE{Testing}
\STATE $\hat{u}_{\text{mean}}, \hat{u}_{\text{var}} \leftarrow $ Macula\big(GenerateROIs\big($\mathbf{o}_{\text{img}}$, MP-Net$(\mathbf{o}_{\text{img}})\big)\big)$
\ENDWHILE
\ENSURE $\hat{u}_{\text{mean}}, \hat{u}_{\text{var}}\text{: control distribution}$
\end{algorithmic}
\end{algorithm}
\end{minipage}
\vspace{-0.2cm}
\end{wrapfigure}

One of the main advantages of our method is that the network can focus on the important/task-related regions of the image, while also eliminating irrelevant parts of the image. This effect can be seen in \cref{fig:MPM}, in which unimportant features (e.g. buildings, sky, trees, etc.) are filtered out with the ROIs.

We resize all \acp{ROI} into the same size as the smallest \ac{ROI}, which is constructed from the farthest focal point generated by the \ac{MP-Net}.
% By this step, with four \acp{ROI}, the concatenated 3D image has half the size of the original input (4x32x32x3 vs. 64x128x3).
The resizing step is inspired by the Glimpse Sensor \citep{glimpse}, where multiple resolution patches were used to improve classification performance. Unlike the Glimpse Sensor, we do not use a simple fully connected layer after the concatenated multi-resolution 2D images. Rather, we process 3D convolutions to extract 3D information among the stacked images.

Through the resizing step, the smallest \ac{ROI} from the farthest focal point maintains its resolution, while the bigger \acp{ROI} downsample to the fixed size, thus obtaining lower resolution. In this manner, the network resembles the parafovea/perifovea area of the macula. In particular, this resemblance emerges from the mechanism that our eyes focus on a specific region with high resolution (fovea), while other surrounding regions are blurred out with lower resolution (parafovea/perifovea).
% thus having a lower resolution (parafovea/perifovea).
% We can think about this resizing step as putting more weights to the smallest and the most important \ac{ROI}.

% how about this kt? do weo need it? Not necessarily. IT's a bit redundant
% i agree. lets kep reading, and maybe we can consider it later. SG
% is this good? what do you think? hold on, im gonna improve it a bit more.

% ok hows' that? Wonderful. ok let me know if you dont like something, and we can think about rephrasing it.  im  sgoruenna continue reading on. SG

% hey kt, i think that paragraph i highlighted is only making things more complicated to understand, but not adding much information to the reader. but you can disagree with it. maybe it is important, in which case we have to explain a bit better. what do you think? isee i see, let me think then. give me a sec.

% Ok. This highlighted part is the only link showing why we came up with the idea of Macula. I'm trying to 'showoff' again that this step is somewhat bio-inspired. Sure.

We stacked images in 3D, resulting in another dimension, z. Because the number of stacked images is relatively small, we do not want this dimension to be reduced and lose information by a pooling layer (\cref{fig:Macula}). Therefore, the 3D max-pooling layers in the network act as 2D max-pooling layers, as they do not pool the z-dimension. For 3D filters, (3, 3, 3) kernels are used.

Finally, we combine MPC and the attention-based image processing, using the \ac{MP-Net} (\cref{fig:MP}) and the Macula-Net (\cref{fig:Macula}), into the \ac{PAPC} algorithm, described in \cref{alg:PAPC}.

\section{Experimental Results}
\label{sec:Experiments}
% \subsection{End-to-End Autonomous Driving with Anomaly Detection}

\begin{wrapfigure}{R}{0.22\textwidth}
\vspace{-2.5cm}
\centering
\subfloat{\includegraphics[width=.22\textwidth]{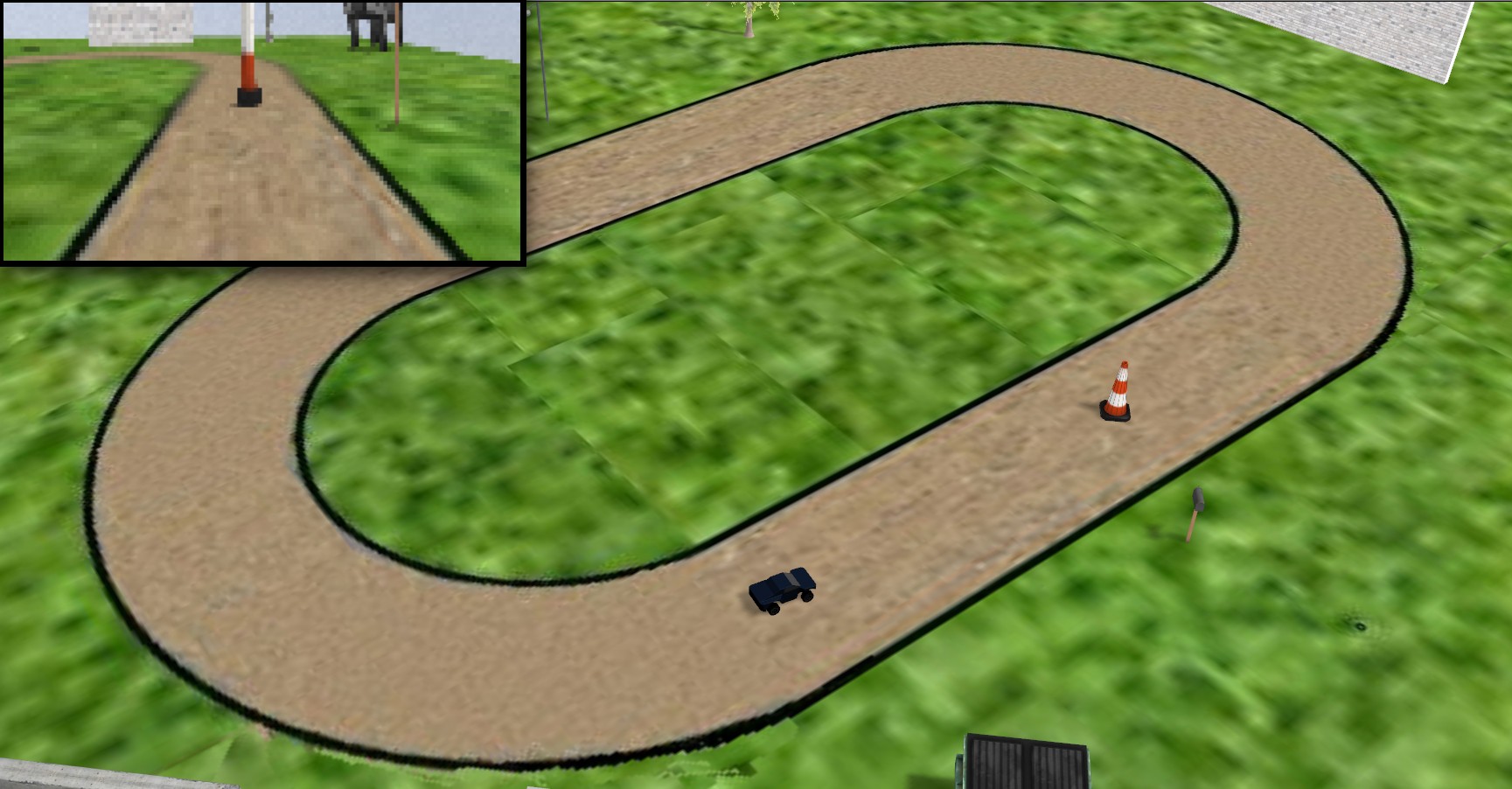}}
\vspace{-3mm}
\subfloat{\includegraphics[width=.22\textwidth]{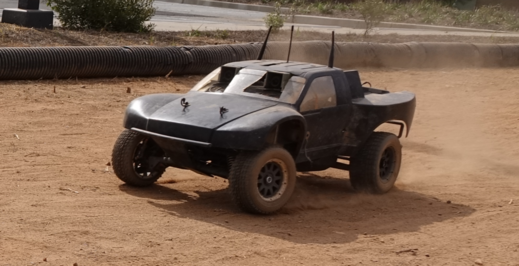}}
% \hfill
% \subfloat{\includegraphics[width=\textwidth]{figs/track.jpg}}
\caption{\footnotesize \textit{Top:} The autonomous driving simulation environment. \textit{Bottom:} The 1/5 scale AutoRally vehicle with onboard cameras used for the experiments.}
% \vspace{1.0cm}
\label{fig:car}
\end{wrapfigure}

For our experiments, we tested our algorithm on autonomous driving tasks in a simulated environment using the ROS Gazebo simulator \citep{ROS} as well as with real hardware with a 1/5 scale AutoRally vehicle \cite{Autorally}. In particular, we show successful results of our attention mechanism for evaluating safety conditions. We obtain these results by examining our model's behavior when encountering safety hazard obstacles in the vehicle's path.

All of the simulation and real-world experiments including the failure cases can be found in the supplementary material and the video.

\subsection{Setup}

\begin{wrapfigure}{R}{0.35\textwidth}
\begin{minipage}{0.35\textwidth}
\vspace{-0.9cm}
\begin{table}[H]
    \scriptsize
     \begin{center}
     \begin{tabular}{ p{0.9cm} p{1.6cm} p{1.0cm} }
     \toprule
     Obstacles & DropoutVGG\citep{Lee2019EarlyFailure} & PAPC[Ours] \\
     \cmidrule(r){1-1}\cmidrule(lr){2-2}\cmidrule(l){3-3}
     \includegraphics[width=0.15\textwidth]{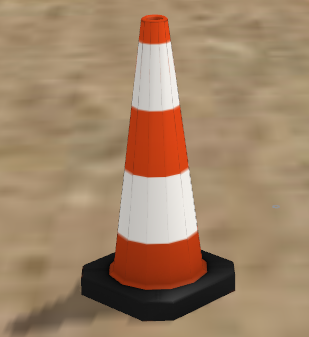}
     &
     \begin{tabular}[b]{@{}c@{}} \multicolumn{1}{r}{Min: 0.37 m} \\ \multicolumn{1}{r}{Avg: 0.39 m} \\
     \multicolumn{1}{r}{Max: 0.42 m} \end{tabular}
     & 
     \begin{tabular}[b]{@{}c@{}} \multicolumn{1}{r}{\textbf{4.28} m} \\ \multicolumn{1}{r}{\textbf{6.87} m} \\
     \multicolumn{1}{r}{\textbf{9.22} m} \end{tabular} \\
     \midrule
     \includegraphics[width=0.15\textwidth]{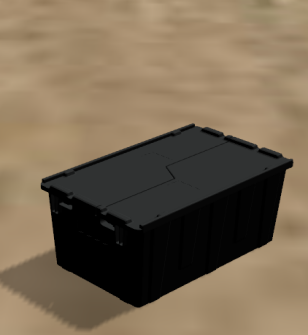}
     &
     \begin{tabular}[b]{@{}c@{}} \multicolumn{1}{r}{Min: 2.20 m} \\ \multicolumn{1}{r}{Avg: 2.54 m} \\
     \multicolumn{1}{r}{Max: 2.86 m} \end{tabular}
     & 
     \begin{tabular}[b]{@{}c@{}} \multicolumn{1}{r}{\textbf{4.81} m} \\ \multicolumn{1}{r}{\textbf{5.48} m} \\
     \multicolumn{1}{r}{\textbf{6.25} m} \end{tabular} \\
     \midrule
     \includegraphics[width=0.15\textwidth]{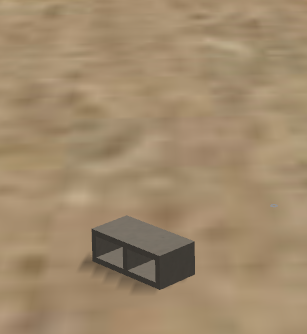}
     &
     \begin{tabular}[b]{@{}c@{}} \multicolumn{1}{r}{Min: 0.00 m} \\ \multicolumn{1}{r}{Avg: 0.00 m} \\
     \multicolumn{1}{r}{Max: 0.00 m} \end{tabular}
     & 
     \begin{tabular}[b]{@{}c@{}} \multicolumn{1}{r}{\textbf{6.80} m} \\ \multicolumn{1}{r}{\textbf{7.25} m} \\
     \multicolumn{1}{r}{\textbf{7.83} m} \end{tabular} \\
     \midrule
     \includegraphics[width=0.15\textwidth]{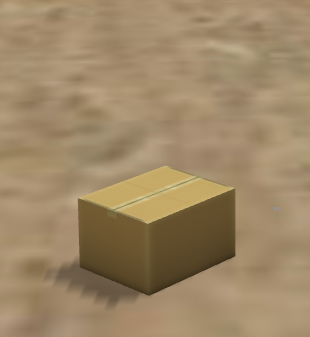}
     &
     \begin{tabular}[b]{@{}c@{}} \multicolumn{1}{r}{Min: 2.12 m} \\ \multicolumn{1}{r}{Avg: 2.25 m} \\
     \multicolumn{1}{r}{Max: 2.44 m} \end{tabular}
     & 
     \begin{tabular}[b]{@{}c@{}} \multicolumn{1}{r}{\textbf{7.62} m} \\ \multicolumn{1}{r}{\textbf{6.87} m} \\
     \multicolumn{1}{r}{\textbf{8.33} m} \end{tabular} \\
     \midrule
     \includegraphics[width=0.15\textwidth]{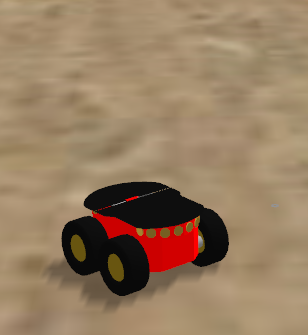}
     &
     \begin{tabular}[b]{@{}c@{}} \multicolumn{1}{r}{Min: 1.28 m} \\ \multicolumn{1}{r}{Avg: 2.06 m} \\
     \multicolumn{1}{r}{Max: 2.44 m} \end{tabular}
     & 
     \begin{tabular}[b]{@{}c@{}} \multicolumn{1}{r}{\textbf{6.55} m} \\ \multicolumn{1}{r}{\textbf{7.51} m} \\
     \multicolumn{1}{r}{\textbf{8.17} m} \end{tabular} \\
     \midrule
     \midrule
     \includegraphics[width=0.15\textwidth]{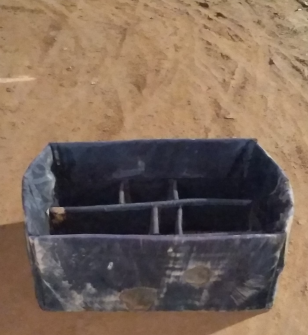}
     &
     \begin{tabular}[b]{@{}c@{}} \multicolumn{1}{r}{Min: 0.00 m} \\ \multicolumn{1}{r}{Avg: 0.63 m} \\
     \multicolumn{1}{r}{Max: 2.51 m} \end{tabular}
     & 
     \begin{tabular}[b]{@{}c@{}} \multicolumn{1}{r}{\textbf{10.58} m} \\ \multicolumn{1}{r}{\textbf{11.28} m} \\
     \multicolumn{1}{r}{\textbf{14.96} m} \end{tabular} \\
     \midrule
     \includegraphics[width=0.15\textwidth]{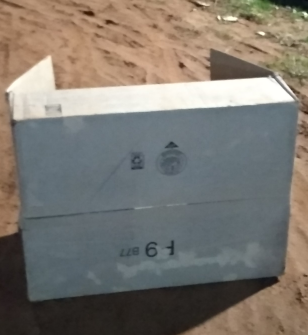}
     &
     \begin{tabular}[b]{@{}c@{}} \multicolumn{1}{r}{Min: 0.00 m} \\ \multicolumn{1}{r}{Avg: 0.26 m} \\
     \multicolumn{1}{r}{Max: 1.29 m} \end{tabular}
     & 
     \begin{tabular}[b]{@{}c@{}} \multicolumn{1}{r}{\textbf{6.91} m} \\ \multicolumn{1}{r}{\textbf{12.55} m} \\
     \multicolumn{1}{r}{\textbf{14.63} m} \end{tabular} \\
     \midrule
     \includegraphics[width=0.15\textwidth]{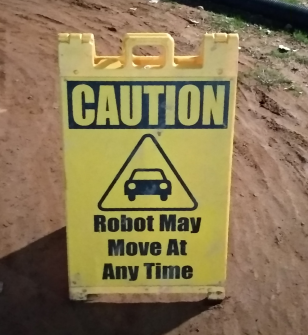}
     &
     \begin{tabular}[b]{@{}c@{}} \multicolumn{1}{r}{Min: 0.00 m} \\ \multicolumn{1}{r}{Avg: 0.67 m} \\
     \multicolumn{1}{r}{Max: 4.11 m} \end{tabular}
     & 
     \begin{tabular}[b]{@{}c@{}} \multicolumn{1}{r}{\textbf{6.17} m} \\ \multicolumn{1}{r}{\textbf{10.09} m} \\
     \multicolumn{1}{r}{\textbf{13.25} m} \end{tabular} \\
     \bottomrule
     \end{tabular}
     \caption{\footnotesize{Distance left from the object when the network detects it. Top 5 objects are tested in the ROS Gazebo simulator and the bottom 3 objects are tested with our real hardware.}}
     \vspace{-1cm}
     \label{tbl:distance_left}
     \end{center}
\end{table}
\end{minipage}
\end{wrapfigure}

We conducted 100 test runs with 5 different obstacles in ROS Gazebo to evaluate \ac{PAPC}'s performance compared to the state of the art. In the real hardware experiments, we conducted 10 trials per obstacle and per network for comparison. All of the experiments were done with NVIDIA GeForce GTX 1050 Ti GPU for the real hardware experiments and 1060 GPU for simulation experiments.

\ac{MP-Net} and Macula-Net were both trained with Adam \citep{adam} optimizer in TensorFlow \citep{tensorflow}. Additionally, we used the Concrete Dropout method \citep{Gal2017Concrete} to find the optimal dropout probability per layer in our Bayesian Network. After every convolution and fully connected layer, we performed batch normalization \citep{batchnorm} to speed up the training and there was no data aggregation involved except for the 3D stack part.
% and all models were trained in batch.

We set a threshold for the output variance signal to 3-10 times larger than the maximum value of the usual variance in the normal scenario (without any novel obstacles). The specific threshold we set depends on the number of samples we choose for the MC-dropout. The usual output variance in the normal situation without any novel object in the scene was between $10^{-5}$ and $10^{-4}$, depending on the number of MC samples as well. We used this threshold to classify an unsafe driving condition and execute an emergency stop.

% First, we run the expert model predictive controller \citep{MPCDDP} to drive the vehicle around an oval track for 100 laps. With the output control and state trajectories from MPC, we train the MP-Net and the Macula-Net as described in previous sections. The control action in our experiments is the vehicle steering command given by a real number in $[-1.0, 1.0]$. 

In addition, we observed that out of the two uncertainties (epistemic and aleatoric) we get from our Bayesian network trained with the \textit{heteroscedastic} loss function $\mathcal{L}_h$, the value of the epistemic uncertainty showed a drastic change reacting to the novel obstacles in the vehicle's trajectory whereas the value of the aleatoric uncertainty does not show such a big difference. This is reasonable because the epistemic uncertainty originates from the lack of data and is therefore expected to provide large variance given input data from the tail of the distribution. Therefore, we used the epistemic uncertainty to quantify the safety of the navigation condition under our network controller.

% The \ac{MP-Net} takes as input full view of images and is trained to predict the planned path by imitating our model predictive controller.
% In addition, we found that the \ac{MP-Net} is trained to map relevant features to the target output, which is the planned trajectories in \ac{MP-Net}. As we can see in the activated feature maps at each layer in \cref{fig:heatmap}, the extracted important features were the line information of the track. In other words, the \ac{MP-Net} is trained to map the track part of the image to the corresponding output (future path in few seconds ahead). As the \ac{MP-Net} gets the full view of the image as input, the new obstacle does not affect/fool the network output most of the time until the obstacle dominates the track in the image. We tested with putting different kinds of objects on the track and getting rid of some features like trees or buildings (in simulation), but neither of them could fool the network because the most important features, track boundaries, still existed in the image. Therefore, we can believe the \ac{MP-Net} to predict the correct trajectories.
\subsection{Result analysis}

Our policy not only performs the original driving task, but it can also assess the safety of the current navigation conditions by quantifying uncertainty in its control policy. In particular, this safety mechanism worked successfully with our vehicle promptly detecting safety hazard obstacles in sufficient time.
\begin{wrapfigure}{R}{0.37\textwidth}
    \vspace{-0.8cm}
    \centering
    \begin{minipage}[t]{0.37\textwidth}
    \includegraphics[width=\textwidth]{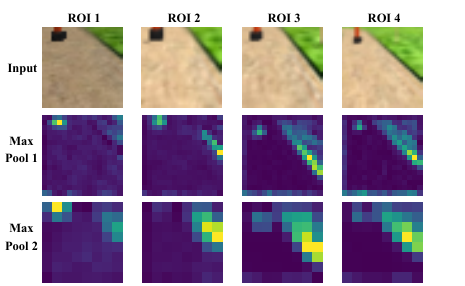}
    \vspace{-0.3cm}
    \caption{\footnotesize{Input stacked 3D ROIs and the averaged feature activation maps after the first two max-pooling layers. Lighter color represents more activation. The smallest ROI, ROI 1, is more activated by the new object than the track/road boundary. As the ROI becomes larger, it tends to be activated from the track boundary than from the new object. }}\label{fig:heatmap}
    \vspace{-0.9cm}
    \end{minipage}
\end{wrapfigure}

To quantify the quality of our safety method enhanced with our attention mechanism, we place novel obstacles in the track and record the distance of the vehicle from the obstacle right when the system declared unsafe driving conditions.

% (\cref{tbl:distance_left}), which was never been seen in the training data on the vehicle's path.
We showed in our previous work \citep{Lee2019EarlyFailure} the increased variance signal from the \ac{BNN} when a vehicle saw a novel object on the road. However, as previously mentioned, the increase of the variance signal was not fast enough to avoid new obstacles at a high speed.

With our results summarized in \cref{tbl:distance_left}, we show that our \ac{PAPC} algorithm has a significantly faster mechanism for identifying unsafe driving conditions. Our method enables the vehicle to stop significantly farther away from the safety hazard obstacles, therefore providing the vehicle with enough time to avoid any collisions.
% To avoid collision of the vehicle driving at around 5 m/s with an obstacle, 1-2 seconds (5-10m) is required for the vehicle to take proper action. Within these 1-2 seconds, a human expert or model predictive controller can take control of the vehicle and avoid the obstacle.
% since the model predictive controllers do optimal path planning and control within 20-50ms.

\subsection{Analysis of attention mechanism}
% Using only the smallest \ac{ROI} around the farthest focal point, it is hard for a network to learn a task since the smallest \ac{ROI} does not contain enough information related to the task. However, even with a low resolution, we have bigger \acp{ROI} which have some information related to given tasks. With a combination of this smallest \ac{ROI} and the largest \ac{ROI}, the model can learn the task as well as focus on important regions of the image to pay attention to and report safety threats.

% We can see that through the \ac{PAPC} algorithm, the input image data to the Macula-Net excludes unimportant features (e.g. buildings, trees, sky, etc.) and only focus on the part where the \ac{MPC} guides the network to focus on.

We analyze the MP-Net by plotting the averaged feature maps after the first two max-pooling layers per ROI (\cref{fig:heatmap}). ROI 1 (fovea) is activated to the feature of a new obstacle, more than the track boundary. However, interestingly, as the ROI becomes larger and downsampled, we can see that the ROI is more activated by the track boundary. From this combination of multi-resolution 3D inputs, the \ac{PAPC} algorithm is able to detect distant obstacles while focusing on the task-related features.

Moreover, we can see that in \cref{fig:heatmap} the deeper layers (closer to the output layer) tend to focus more on a single feature. For example in ROI 2, after the first max-pooling layer, the neurons are activated almost equally from both the new object and the track boundary. However, after two convolutional layers and a max-pooling layer, the neurons are activated only from the feature of the track boundary. This pattern is also seen in other \acp{ROI}.

Finally, in \cref{fig:heatmap}, we can also see that the resolution of the ROIs becomes lower as the \ac{ROI} becomes larger since it needs more downsampling. From this downsampling, we can observe the loss of information on the bigger \acp{ROI}, where the brightness also has been changed.

% \section{Discussion}
% \label{sec:discussion}
% For our future work, we would like to explore smaller network structures instead of the VGG-based network. We were able to run our robot in real-time (15-20Hz), but we could only sample around 25 samples from the Monte Carlo dropout. We also tested with more samples in lower control frequency and saw less noisy anomaly detection signal from the output variance of our Bayesian Network.

% While  PAPC is able to detect novel objects and increase the output predicted uncertainty, the duration of this detection is instantaneous. This because the detection depends on the time interval during which the ROI with the highest resolution on the tip of the MP-Net predicted trajectory overlaps with the new object.  In future work, we plan to combine tracking mechanisms with \ac{PAPC} so that to achieve an increase in predicted uncertainty of the Macula-Net for as long as the new object is in the field of view of the vehicle.  

\section{Conclusion} 
\label{sec:conclusion}
In this work, we have presented a novel IPA, namely the \ac{PAPC} algorithm, which performs vision-based navigation and incorporates a highly proficient attention mechanism to assess the safety of the navigation conditions under the network controller. We emphasize here that the PAPC can be used in any autonomous system that performs navigation using visual sensors for safe path planning and control (e.g. visuomotor for manipulation \citep{LevineVisuomotor}).

In addition, while our initial goal is to use the \ac{PAPC} architecture as the main system for navigation, its operational role can also be used as a secondary safety controller with the sole purpose of evaluating the safety of the navigation conditions. \ac{PAPC}'s safety evaluation method can be integrated into decision-making modules to build other robust/adaptive controllers.
% Furthermore, this anomaly detection approach can be used for improving data aggregation during learning or for intelligent exploration in unknown or partially known environments.

The proposed algorithm is validated in both simulation and real-world by outperforming the state-of-the-art approach in a safety-aware autonomous driving task.

% In this work, we view perceptual control policies as Information  Processing Architectures, or IPAs for short, and propose a new architecture to support an algorithm for perceptual attention-based control to perform an early detection of new objects using a camera.
% The architecture for \ac{PAPC} consists primarily of two CNNs, namely the \ac{MP-Net} and the Macula-Net which is introduced for the first time in this paper.
% The \ac{MP-Net} is trained so that to generate predicted state trajectories using vision. These trajectories determine ROIs in the image with variable resolution.
% Using as input the aforementioned multi-resolution ROIs, the Macula-Net is able to focus on relevant, with respect to the task, areas of the input visual information, detect novel objects and control the vehicle in consideration. We validated our proposed attention-based deep Bayesian network in both ROS Gazebo simulator and the real hardware for a safety-aware autonomous driving task. \ac{PAPC} was able to detect different obstacles quickly in comparison with the state-of-the-art approach of the end-to-end Bayesian network \citep{Lee2019EarlyFailure}.

\acknowledgments{\
This work was supported by Amazon Web Services (AWS) and Komatsu Ltd.}

\bibliography{PAPC}

\begin{thebibliography}{26}
\providecommand{\natexlab}[1]{#1}
\providecommand{\url}[1]{\texttt{#1}}
\expandafter\ifx\csname urlstyle\endcsname\relax
  \providecommand{\doi}[1]{doi: #1}\else
  \providecommand{\doi}{doi: \begingroup \urlstyle{rm}\Url}\fi

\bibitem[Shelhamer et~al.(2017)Shelhamer, Long, and Darrell]{fcn}
E.~Shelhamer, J.~Long, and T.~Darrell.
\newblock \href{https://doi.org/10.1109/TPAMI.2016.2572683}{Fully Convolutional
  Networks for Semantic Segmentation}.
\newblock \emph{IEEE Transactions on Pattern Analysis and Machine
  Intelligence}, 39\penalty0 (4):\penalty0 640--651, Apr. 2017.
\newblock ISSN 0162-8828.

\bibitem[Ren et~al.(2015)Ren, He, Girshick, and Sun]{fasterrcnn}
S.~Ren, K.~He, R.~Girshick, and J.~Sun.
\newblock \href{https://arxiv.org/abs/1506.01497}{Faster R-CNN: Towards
  Real-Time Object Detection with Region Proposal Networks}.
\newblock In C.~Cortes, N.~D. Lawrence, D.~D. Lee, M.~Sugiyama, and R.~Garnett,
  editors, \emph{Advances in Neural Information Processing Systems 28}, pages
  91--99. Curran Associates, Inc., 2015.

\bibitem[He et~al.(2017)He, Gkioxari, Dollar, and Girshick]{maskrcnn}
K.~He, G.~Gkioxari, P.~Dollar, and R.~Girshick.
\newblock
  \href{http://openaccess.thecvf.com/content_ICCV_2017/papers/He_Mask_R-CNN_ICCV_2017_paper.pdf}{Mask
  R-CNN}.
\newblock In \emph{IEEE International Conference on Computer Vision ({ICCV})},
  Oct 2017.

\bibitem[Bojarski et~al.(2016)Bojarski, {Del Testa}, Dworakowski, Firner,
  Flepp, Goyal, Jackel, Monfort, Muller, Zhang, Zhang, Zhao, and
  Zieba]{bojarski2016end}
M.~Bojarski, D.~{Del Testa}, D.~Dworakowski, B.~Firner, B.~Flepp, P.~Goyal,
  L.~D. Jackel, M.~Monfort, U.~Muller, J.~Zhang, X.~Zhang, J.~Zhao, and
  K.~Zieba.
\newblock \href{https://arxiv.org/abs/1604.07316}{End to End Learning for
  Self-Driving Cars}.
\newblock \emph{arXiv}, Apr 2016.

\bibitem[Pan et~al.(2018)Pan, Cheng, Saigol, Lee, Yan, Theodorou, and
  Boots]{PanRSS18}
Y.~Pan, C.-A. Cheng, K.~Saigol, K.~Lee, X.~Yan, E.~A. Theodorou, and B.~Boots.
\newblock \href{http://www.roboticsproceedings.org/rss14/p56.pdf}{Agile
  Autonomous Driving using End-to-End Deep Imitation Learning}.
\newblock \emph{Robotics: Science and Systems}, 2018.

\bibitem[Ross et~al.(2011)Ross, Gordon, and Bagnell]{DAgger}
S.~Ross, G.~J. Gordon, and J.~A. Bagnell.
\newblock \href{http://proceedings.mlr.press/v15/ross11a/ross11a.pdf}{A
  Reduction of Imitation Learning and Structured Prediction to No-Regret Online
  Learning}.
\newblock In \emph{Proceedings of the 14th International Conference on
  Artificial Intelligence and Statistics}, volume~15 of \emph{JMLR}, Fort
  Lauderdale, FL, USA, 2011.

\bibitem[{Lee} et~al.(2019){Lee}, {Saigol}, and
  {Theodorou}]{Lee2019EarlyFailure}
K.~{Lee}, K.~{Saigol}, and E.~A. {Theodorou}.
\newblock \href{10.1109/ICRA.2019.8794189}{Early Failure Detection of Deep
  End-to-End Control Policy by Reinforcement Learning}.
\newblock In \emph{2019 International Conference on Robotics and Automation
  (ICRA)}, pages 8543--8549, May 2019.

\bibitem[Lee et~al.(2019)Lee, Wang, Vlahov, Brar, and
  Theodorou]{Lee2019Ensemble}
K.~Lee, Z.~Wang, B.~I. Vlahov, H.~K. Brar, and E.~A. Theodorou.
\newblock \href{https://arxiv.org/abs/1811.12555}{Ensemble Bayesian Decision
  Making with Redundant Deep Perceptual Control Policies}.
\newblock \emph{The 18th IEEE International Conference on Machine Learning and
  Applications (ICMLA)}, 2019.

\bibitem[Tassa et~al.(2008)Tassa, Erez, and Smart]{MPCDDP}
Y.~Tassa, T.~Erez, and W.~D. Smart.
\newblock
  \href{http://papers.nips.cc/paper/3297-receding-horizon-differential-dynamic-programming.pdf}{Receding
  Horizon Differential Dynamic Programming}.
\newblock \emph{Advances in Neural Information Processing Systems 20}, pages
  1465--1472, 2008.

\bibitem[Williams et~al.(2016)Williams, Drews, Goldfain, Rehg, and
  Theodorou]{mppi}
G.~Williams, P.~Drews, B.~Goldfain, J.~M. Rehg, and E.~A. Theodorou.
\newblock
  \href{https://ieeexplore.ieee.org/abstract/document/7487277}{Aggressive
  driving with model predictive path integral control}.
\newblock \emph{2016 IEEE International Conference on Robotics and Automation
  (ICRA)}, 2016.

\bibitem[Blundell et~al.(2015)Blundell, Cornebise, Kavukcuoglu, and
  Wierstra]{pmlr-v37-blundell15}
C.~Blundell, J.~Cornebise, K.~Kavukcuoglu, and D.~Wierstra.
\newblock \href{http://proceedings.mlr.press/v37/blundell15.pdf}{Weight
  Uncertainty in Neural Network}.
\newblock In F.~Bach and D.~Blei, editors, \emph{Proceedings of the 32nd
  International Conference on Machine Learning}, volume~37 of \emph{Proceedings
  of Machine Learning Research}, pages 1613--1622, Lille, France, 07--09 Jul
  2015. PMLR.

\bibitem[Gal and Ghahramani(2016)]{pmlr-v48-gal16}
Y.~Gal and Z.~Ghahramani.
\newblock \href{http://proceedings.mlr.press/v48/gal16.pdf}{Dropout as a
  Bayesian Approximation: Representing Model Uncertainty in Deep Learning}.
\newblock In \emph{Proceedings of The 33rd International Conference on Machine
  Learning}, volume~48 of \emph{Proceedings of Machine Learning Research},
  pages 1050--1059, New York, New York, USA, 20--22 Jun 2016. PMLR.

\bibitem[Lakshminarayanan et~al.(2017)Lakshminarayanan, Pritzel, and
  Blundell]{Lakshminarayanan_17_deepensemble}
B.~Lakshminarayanan, A.~Pritzel, and C.~Blundell.
\newblock
  \href{http://papers.nips.cc/paper/7219-simple-and-scalable-predictive-uncertainty-estimation-using-deep-ensembles.pdf}{Simple
  and Scalable Predictive Uncertainty Estimation using Deep Ensembles}.
\newblock In I.~Guyon, U.~V. Luxburg, S.~Bengio, H.~Wallach, R.~Fergus,
  S.~Vishwanathan, and R.~Garnett, editors, \emph{Advances in Neural
  Information Processing Systems 30}, pages 6402--6413. Curran Associates,
  Inc., 2017.

\bibitem[Kendall and Gal(2017)]{WhatUncertainties}
A.~Kendall and Y.~Gal.
\newblock \href{https://arxiv.org/abs/1703.04977}{What Uncertainties Do We Need
  in Bayesian Deep Learning for Computer Vision?}
\newblock In I.~Guyon, U.~V. Luxburg, S.~Bengio, H.~Wallach, R.~Fergus,
  S.~Vishwanathan, and R.~Garnett, editors, \emph{Advances in Neural
  Information Processing Systems 30}, pages 5574--5584. Curran Associates,
  Inc., 2017.

\bibitem[Boor(1970)]{Bspline}
C.~D. Boor.
\newblock
  \href{https://www.sciencedirect.com/science/article/pii/0021904572900809}{On
  calculating with B-splines}.
\newblock \emph{Journal of Approximation Theory}, 6, 1970.

\bibitem[Boor(1977)]{cdboorformula}
C.~D. Boor.
\newblock
  \href{https://www.jstor.org/stable/2156696?seq=1#page_scan_tab_contents}{Package
  for calculating with B-splines}.
\newblock \emph{SIAM Journal on Numerical Analysis}, 14:\penalty0 441--472,
  1977.

\bibitem[Trucco and Verri(1998)]{coordtransformbook}
E.~Trucco and A.~Verri.
\newblock \emph{Introductory Techniques for 3-D Computer Vision}.
\newblock Prentice Hall PTR, Upper Saddle River, NJ, USA, 1998.
\newblock ISBN 0132611082.

\bibitem[Simonyan and Zisserman(2015)]{VGG}
K.~Simonyan and A.~Zisserman.
\newblock \href{http://arxiv.org/abs/1409.1556}{Very Deep Convolutional
  Networks for Large-Scale Image Recognition}.
\newblock In \emph{International Conference on Learning Representations}, 2015.

\bibitem[Mnih et~al.(2014)Mnih, Heess, Graves, and kavukcuoglu]{glimpse}
V.~Mnih, N.~Heess, A.~Graves, and k.~kavukcuoglu.
\newblock
  \href{http://papers.nips.cc/paper/5542-recurrent-models-of-visual-attention.pdf}{Recurrent
  Models of Visual Attention}.
\newblock In Z.~Ghahramani, M.~Welling, C.~Cortes, N.~D. Lawrence, and K.~Q.
  Weinberger, editors, \emph{Advances in Neural Information Processing Systems
  27}, pages 2204--2212. Curran Associates, Inc., 2014.

\bibitem[Quigley et~al.(2009)Quigley, Conley, Gerkey, Faust, Foote, Leibs,
  Wheeler, and Ng]{ROS}
M.~Quigley, K.~Conley, B.~P. Gerkey, J.~Faust, T.~Foote, J.~Leibs, R.~Wheeler,
  and A.~Y. Ng.
\newblock
  \href{http://www.willowgarage.com/sites/default/files/icraoss09-ROS.pdf}{ROS:
  an open-source Robot Operating System}.
\newblock In \emph{ICRA Workshop on Open Source Software}, 2009.

\bibitem[Goldfain et~al.(2019)Goldfain, Drews, You, Barulic, Velev, Tsiotras,
  and Rehg]{Autorally}
B.~Goldfain, P.~Drews, C.~You, M.~Barulic, O.~Velev, P.~Tsiotras, and J.~M.
  Rehg.
\newblock \href{https://ieeexplore.ieee.org/document/8616931}{AutoRally: An
  Open Platform for Aggressive Autonomous Driving}.
\newblock \emph{IEEE Control Systems Magazine}, 39\penalty0 (1):\penalty0
  26--55, Feb 2019.

\bibitem[Kingma and Ba(2014)]{adam}
D.~P. Kingma and J.~Ba.
\newblock \href{http://arxiv.org/abs/1412.6980}{Adam: {A} Method for Stochastic
  Optimization}.
\newblock \emph{Proceedings of the 3rd International Conference on Learning
  Representations (ICLR)}, abs/1412.6980, 2014.

\bibitem[Abadi et~al.(2015)Abadi, Agarwal, Barham, Brevdo, Chen, Citro,
  Corrado, Davis, Dean, Devin, Ghemawat, Goodfellow, Harp, Irving, Isard, Jia,
  Jozefowicz, Kaiser, Kudlur, Levenberg, Man\'{e}, Monga, Moore, Murray, Olah,
  Schuster, Shlens, Steiner, Sutskever, Talwar, Tucker, Vanhoucke, Vasudevan,
  Vi\'{e}gas, Vinyals, Warden, Wattenberg, Wicke, Yu, and Zheng]{tensorflow}
M.~Abadi, A.~Agarwal, P.~Barham, E.~Brevdo, Z.~Chen, C.~Citro, G.~S. Corrado,
  A.~Davis, J.~Dean, M.~Devin, S.~Ghemawat, I.~Goodfellow, A.~Harp, G.~Irving,
  M.~Isard, Y.~Jia, R.~Jozefowicz, L.~Kaiser, M.~Kudlur, J.~Levenberg,
  D.~Man\'{e}, R.~Monga, S.~Moore, D.~Murray, C.~Olah, M.~Schuster, J.~Shlens,
  B.~Steiner, I.~Sutskever, K.~Talwar, P.~Tucker, V.~Vanhoucke, V.~Vasudevan,
  F.~Vi\'{e}gas, O.~Vinyals, P.~Warden, M.~Wattenberg, M.~Wicke, Y.~Yu, and
  X.~Zheng.
\newblock \href{http://tensorflow.org/}{{TensorFlow}: Large-Scale Machine
  Learning on Heterogeneous Systems}, 2015.
\newblock Software available from tensorflow.org.

\bibitem[{Yarin Gal and Jiri Hron and Alex Kendall}(2017)]{Gal2017Concrete}
{Yarin Gal and Jiri Hron and Alex Kendall}.
\newblock \href{https://arxiv.org/abs/1705.07832}{Concrete Dropout}.
\newblock In \emph{Advances in Neural Information Processing Systems 30}, 2017.

\bibitem[Ioffe and Szegedy(2015)]{batchnorm}
S.~Ioffe and C.~Szegedy.
\newblock \href{http://jmlr.org/proceedings/papers/v37/ioffe15.pdf}{Batch
  Normalization: Accelerating Deep Network Training by Reducing Internal
  Covariate Shift}.
\newblock \emph{Proceedings of the 32nd International Conference on Machine
  Learning}, 37:\penalty0 448--456, 2015.

\bibitem[Levine et~al.(2016)Levine, Finn, Darrell, and
  Abbeel]{LevineVisuomotor}
S.~Levine, C.~Finn, T.~Darrell, and P.~Abbeel.
\newblock \href{http://jmlr.org/papers/v17/15-522.html}{End-to-End Training of
  Deep Visuomotor Policies}.
\newblock \emph{Journal of Machine Learning Research}, 17\penalty0
  (39):\penalty0 1--40, 2016.

\end{thebibliography}
\clearpage
\section*{Appendix}
\renewcommand{\thesubsection}{A\arabic{subsection}}
\subsection{Comparison Plot}
\begin{figure}[h]
\centering
  \begin{minipage}[b]{0.8\textwidth}
    \includegraphics[width=\textwidth]{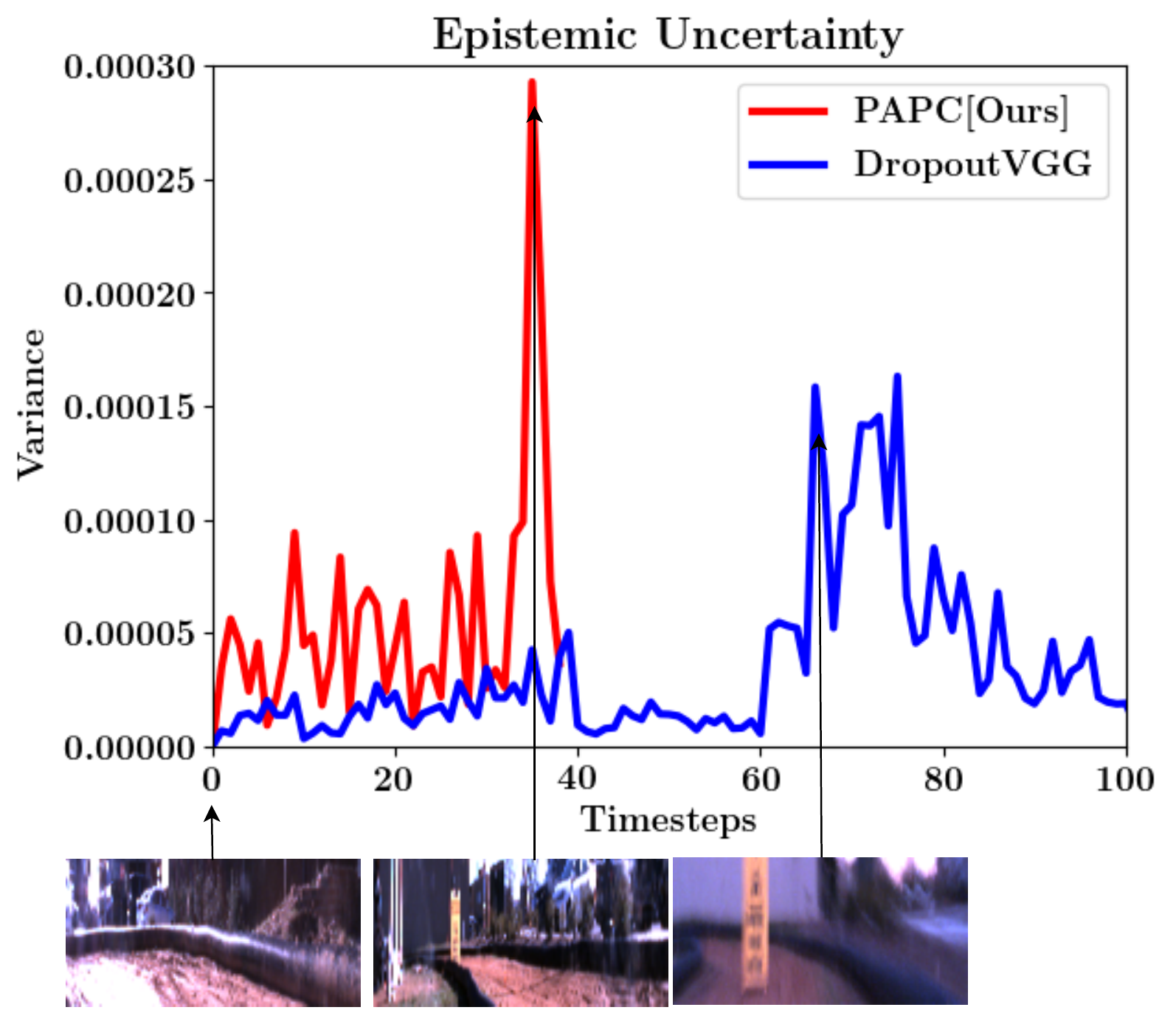}
    \caption{\footnotesize The Output variance plot from Bayesian networks. It shows the epistemic uncertainty from the MC-dropout. \ac{PAPC} shows an earlier detection of the novel obstacle compared to the DropoutVGG \citep{Lee2019EarlyFailure}.} \label{fig:spike}
  \end{minipage}
  \hfill
\end{figure}

\cref{fig:spike} shows the comparison between our proposed approach, PAPC, and the state-of-the-art method DropoutVGG \citep{Lee2019EarlyFailure} for detecting a new obstacle.

\subsection{Failure Cases}

\begin{figure}
\vspace{-0.5cm}
  \centering
    \includegraphics[width=0.5\textwidth]{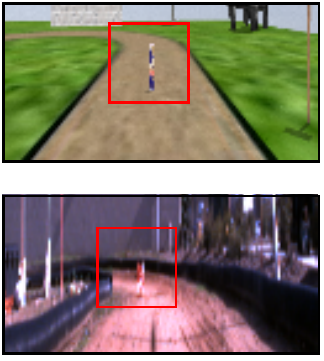}
    \caption{\footnotesize The failure cases of detecting a novel beer can in the ROS Gazebo simulator and the detergent container at the real track. Even though the fovea (red box) caught the objects, the \ac{PAPC} did not output any meaningful signal.}\label{fig:failure}
\end{figure}

As shown in \cref{fig:failure}, our network sometimes fails, depending on the object size or color. \ac{PAPC} was not able to detect a can in the simulation environment and a detergent container in the real world. We argue that their distributions were not different enough from the training data, even though the fovea \ac{ROI} caught the object correctly. However, when the vehicle gets closer to the objects, the fovea ROI has already passed it, so no more strong attentions exist at that time. We believe these kinds of smaller objects or those having a similar distribution to the training data without obstacles can be detected by increasing the number of focal points and ROIs, with the fovea having a smaller window. Having more \acp{ROI} will require faster GPUs and smaller network structures to run the network in real time, however.

\subsection{Data Distribution Visualization with t-SNE} \label{sec:tsne}

\begin{figure}
    \centering
    \includegraphics[width=\textwidth]{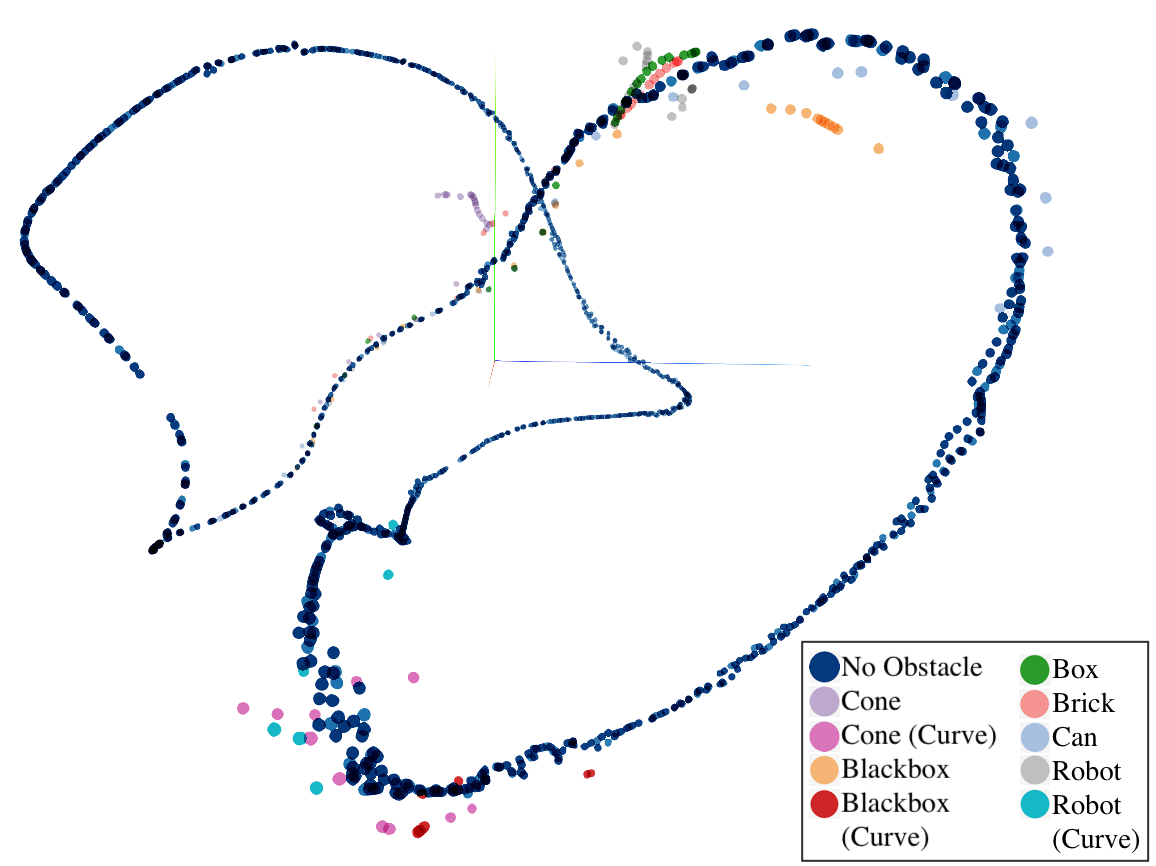}
    \caption{\footnotesize{A screenshot of the 3D t-SNE plot of the 3rd max-pooling layer in the Macula-Net. The colorful cluster shows how the Macula-Net is able to abstract the presence of an obstacle in its way. The 3D plot is best viewed in the video: \url{https://youtu.be/-Zmi0HCvM9I}.}}\label{fig:tsne}
    \vspace{-0.5cm}
\end{figure}

Because the Macula-Network uses the Bayesian dropout \citep{pmlr-v48-gal16} approach to determine whether ROI contains a new obstacle, it is useful to analyze how the distribution of the ROIs with and without obstacle differ from one another. We use the t-SNE technique for dimensionality reduction in order to visualize the high dimensional \ac{ROI} images that the Macula-Network takes in as input (\cref{fig:tsne}). We run t-SNE with perplexity values around 60 with simulated and real-world ROI data. In addition, we not only run t-SNE with the ROI images but also with the output of the first fully-connected layer in the Macula-Net. Using a middle layer allows us to visualize how the Macula-Net itself interprets the ROI images. In \cref{fig:heatmap}, we show that the Macula-Net features of the images with obstacles lie in a noticeably different distribution (colorful cluster) as the images with no obstacles. This shows that the Macula-Net was able to capture a change in the distribution of the ROIs when an obstacle is put in the track. Because these image samples with obstacles lie in a different distribution as compared to the training data, the Bayesian network inside the Macula-Net will output control values with high variance, thus indicating the policy that it is no longer safe to drive. In this manner, the algorithm will be able to tell when to concede control to a safer controller.

\subsection{Expert Model Predictive Controller used to train the model} \label{sec:expert}
In this work, \acl{iLQG/MPC-DDP} \citep{MPCDDP} was used as an expert controller to drive the vehicle on the track. Using an onboard GPS and a IMU sensor, the state estimation was performed and the particle filter was used for filtering it. With the estimated 12 states $x=[U, V, W, \phi, \theta, \psi, \dot{U}, \dot{V}, \dot{W}, \dot{\phi}, \dot{\theta}, \dot{\psi}]$, the task-dependent cost function \cref{eq:optcontrolprob} was optimized with the nonlinear vehicle dynamics model in the DDP algorithm. In our experiments, the nonlinear vehicle dynamics are learned from collected data using deep feedforward neural networks. DDP optimization results then provided planned future path of the model within a time horizon along with control trajectories. DDP was performed iteratively with a receding-horizon fashion in 50Hz.

For autonomous driving task on a track, the running cost and the terminal cost was designed to stay at the center of the track depending on $U$ and $V$. Before running DDP, the center line data $(U, V)$ of the track/map were collected with GPS, and used as some notion of way points which our robot could follow. Additionally, to maintain a certain speed, a quadratic speed cost depending on $\dot{U}$ and $\dot{V}$ was added to the running cost and the terminal cost.

\end{document}